\documentclass[twoside]{article}
\usepackage[a4paper]{geometry}
\usepackage[utf8]{inputenc}
\usepackage{RR}
\usepackage{hyperref}
\usepackage[american]{babel} 
\RRNo{8563}
\RRdate{July 2014}
\RRauthor{
Jérémie Mary
\and
Romaric Gaudel
\and
Philippe Preux
  \thanks[sfn]{Université de Lille, LIFL (UMR CNRS), INRIA, Villeneuve d'Ascq, France, \url{firstname.lastname@univ-lille3.fr}}%
}
\authorhead{Mary \& Gaudel \& Preux}
\RRtitle{Démarrage à froid des systèmes de recommendation par bandits}
\RRetitle{Bandits Warm-up Cold Recommender Systems}
\titlehead{Bandits Warm-up Cold Recommender Systems}
\RRresume{
  Nous nous intéressons au problème du démarrage à froid dans les systèmes de recommendation. Nous supposons ne disposer d'aucune information, que ce soit à propos des utilisateurs ou des produits. Nous considérons le cas où nous n'avons accès qu'à un ensemble de notes données à des produits par des utilisateurs. La plupart des travaux concernant ce problème considèrent une approche par lôts et utilisent la validation croisée pour régler les paramètres. La méthode classique consiste à réaliser une décomposition de faible rang de la matrice des notes en minimisant l'erreur quadratique moyenne sur un sous-ensemble des notes disponibles. Cette factorisation est interprêtée comme exhibant des facteurs latents décrivant les produits et les utilisateurs. Dans ce rapport, notre contribution concerne 5 points. Tout d'abord, nous explicitons les problèmes posés par ce type d'approches par lôt pour des utilisateurs ou des produits ayant très peu de notes qui leur sont associées (utilisateurs et produits froids). Ensuite, nous proposons une approche séquentielle qui se rapproche fortement du mode d'utilisation réelle des systèmes de recommendation. Cette approche est inspirée par le problème du bandit multi-bras. Cette méthodologie permet de transformer tout jeu de données issu d'un système de recommendation (tels Netflix, MovieLens, ...) en un jeu de données séquentiel. Alors, nous explicitons une forte connexion entre les bandits contextuels et la factorisation de matrices~; nous pensons que la mise à jour de cette relation est la contribution conceptuelle essentielle de ce rapport~; cette relation éclaire cette problématique d'un jour nouveau. Cela nous amène à un nouvel algorithme qui prend en charge le dilemme exploration/exploitation existant dans le problème du démarrage à froid. Finalement, une étude expérimentale de cet algorithme montre que l'approche fonctionne efficacement pour gérer le démarrage à froid sur des jeux de données disponibles publiquement. Pour résumer nos contributions en une phrase, l'objectif de ce rapport est de mettre à jour un pont entre les systèmes de recommendation basés sur la factorisation de matrices d'une part, les bandits contextuels d'autre part.
}
\RRabstract{
  We address the cold start problem in recommendation systems assuming
  no contextual information is available neither about users, nor
  items. We consider the case in which we only have access to a set of
  ratings of items by users. Most of the existing works consider a
  batch setting, and use cross-validation to tune parameters. The
  classical method consists in minimizing the root mean square error
  over a training subset of the ratings which provides a factorization
  of the matrix of ratings, interpreted as a latent representation of
  items and users. Our contribution in this paper is 5-fold. First, we
  explicit the issues raised by this kind of batch setting for users
  or items with very few ratings. Then, we propose an online setting
  closer to the actual use of recommender systems; this setting is
  inspired by the bandit framework. The proposed methodology can be
  used to turn any recommender system dataset (such as Netflix,
  MovieLens,...) into a sequential dataset. Then, we explicit a strong
  and insightful link between contextual bandit algorithms and matrix
  factorization; this leads us to a new algorithm that tackles the
  exploration/exploitation dilemma associated to the cold start
  problem in a strikingly new perspective. Finally, experimental
  evidence confirm that our algorithm is effective in dealing with the
  cold start problem on publicly available datasets. Overall, the goal
  of this paper is to bridge the gap between recommender systems based
  on matrix factorizations and those based on contextual bandits.
}
\RRmotcle{Système de recommendation, démarrage à froid, factorisation de matrice, facteurs latents, bandit multi-bras, bandit contextuel, dilemme exploration/exploitation, prise de décision séquentielle dans l'incertain, algorithme efficace.}
\RRkeyword{Recommendation system, cold-start problem, matrix factorization, latent factors, multi-armed bandit, contextual bandit, exploration/exploitation dilemma, sequential decision making under uncertainty, efficient algorithm.}
\RRprojet{SequeL}
\RCLille 

\usepackage{amssymb}
\usepackage{graphicx}

\usepackage[utf8]{inputenc}
\usepackage[american]{babel}
\usepackage{subfigure}

\usepackage{algorithm}
\usepackage{algorithmic}

\newcommand{\icmlVersion}[1]{}

\usepackage{url}
\urldef{\mails}\path|{jeremie.mary,romaric.gaudel,philippe.preux}@univ-lille3.fr|

\renewcommand{\algorithmicrequire}{\textbf{\quad~~~Input:}}
\renewcommand{\algorithmicensure}{\textbf{\quad~~~Input/Output:}}
\usepackage{amsfonts,amsmath}

\newcommand{\UCBonAllUsers}{UCB.on.all.users}
\newcommand{\GreedyALSWR}{Greedy.ALS-WR}
\newcommand{\GreedyALS}{Greedy.ALS}

\usepackage{array}

\usepackage[usenames,dvipsnames]{xcolor}

\usepackage{xspace}
\newcommand{\ie}{\textit{i.e.}\@\xspace}
\newcommand{\etal}{\textit{et al.}\@\xspace}
\newcommand{\knownRatings}{{\cal S}}
\newcommand{\rs}{RS}

\graphicspath{{figures/}}

\usepackage{tikz}
\usetikzlibrary{arrows,snakes,shapes}
\newcommand{\vect}[1]{\mathbf{#1}}
\newcommand{\mat}[1]{\mathbf{#1}}

\renewcommand{\epsilon}{\varepsilon}

\newcommand{\A}{{\mat A}}
\newcommand{\B}{{\mat B}}

\newcommand{\Id}{{\mat{Id}}}

\newcommand{\J}{{\cal J}}

\newcommand{\R}{\mathbb{R}}
\newcommand{\Rknown}{{\mat R}}

\newcommand{\U}{{\mat U}}
\newcommand{\V}{{\mat V}}

\newcommand{\vv}{{\vect v}}
\newcommand{\uu}{{\vect u}}

\DeclareMathOperator*{\argmax}{argmax}
\DeclareMathOperator*{\argmin}{argmin}

\DeclareMathOperator*{\MF}{Matrix Factorization}

\def\beglab{\begin{equation} \label}
\def\endlab{\end{equation}}
\def\beglabc{\begin{equation*} }
\def\endlabc{\end{equation*}}

\DeclareMathOperator{\cumrew}{CumRew}
\DeclareMathOperator{\cumreg}{Regret}

\newcommand{\expexp}{\texttt{Explore-Exploit}}
\newcommand{\epsgreedy}{$\epsilon$-greedy}

\newcommand{\keepForLongVersionIfAny}[1]{}

\usepackage[citestyle=authoryear,backend=bibtex]{biblatex}
\begin{document}
\makeRR

\section{Introduction}


We consider the online version of the problem of the recommendation of
items, that is, the one faced by websites. Items may be ads, news,
music, videos, movies, books, diapers, ... Daily, of even more often,
these systems have to cope with users that have never visited the
website, and new items introduced in the catalog. Appetence of the new
users towards available items, and appeal of new items towards
existing users have to be estimated as fast as possible: this is the
cold start problem.  Currently, this situation is handled thanks to
side information available either about the user, or about the item
(see \cite{DBLP:conf/nips/AgarwalCEMPRRZ08,contextualRecommendation}).
In this paper, we consider this problem from a different
perspective. Though perfectly aware of the potential utility of side
information, we consider the problem without any side information,
only focussing on acquiring appetence of new users and appeal new
items as fast as possible; side information can be mixed with the
ideas presented in this paper. This combination is left as future
work. This poblem fits perfectly into the sequential decision making
framework, and more specifically, the bandit without side information
setting. However, in rather sharp contrast with the traditional bandit
setting, here the set of bandits is continuously being renewed; the
number of bandits is not small, though not being huge (from a few
dozens to hundreds arms in general, up to dozens of millions in some
application): this makes the problem very different from the 2-armed
bandit problem, though asymptotic approximation is still irrelevant;
we look for efficient and effective ways to achieve this goal, since
we want the proposed solution to be able to cope with real
applications on the web. For obvious practical and economical reasons
for real applications, the strategy can not merely consist in
repeatedly presenting all available items to users until the appetence
seems accurately estimated. We have to consider the problem as an
exploration vs.\@ exploitation problem in which exploration is a
necessary evil to acquire information and eventually improve the
performance of the recommendation system (RS for short).

This being said, comes the problem of the objective function to
optimize. Since the Netflix challenge, at least in the machine
learning community, the recommendation problem is often boiled down
to a matrix factorization problem, performed in batch, learning on a
training set, and minimizing the root mean squared error (RMSE) on a
testing set. However, the RMSE comes along very heavy flaws:
\begin{itemize}
  \item Using the RMSE makes no difference between the items that are highly rated by a user and items poorly rated by the same user; however, for a user, there is a big difference between well rated items and the others: the user wants to be recommended with items she will rate high; she does not care about unattractive items; to illustrate that idea in a rating context alike the Netflix challenge using integers in the range 1 to 5, making an error between a 4 and a 5 is qualitatively very different from making an error between 1 and 2. Furthermore, the restricted set of possible ratings implies that a 5 corresponds to more or less highly rated items. If ratings were real numbers, 5 would spread into $[4.5, 5.5]$ allowing a more precise ranking of preferences by each user. Finally, it is well-known that users have a propensity to rate items they like, rather than rate items they dislike \cite{steck:kdd2010}.

  \item RMSE does not make any difference between the outcome of
    recommending an item to a heavy user (a user who has already rated
    a lot of items) as to observe the outcome of the first
    recommendation to a user during her first visit to the website.

  \item Usually, the training set and the testing set are unordered,
    all information regarding the history of the interactions being
    left aside. Then, we consider average appetence over time,
    completely neglecting the fact that a given item does not have the
    same appeal from its birth to its death, and the fact that the
    appeal of items is often correlated to the set of available items
    at a given time, and those available in the past. \cite{koren:td} has shown the importance of taking timestamps into account.

  \item Though item recommendation is often presented as a prediction problem, it is really a ranking problem: however, RMSE is not meant to evaluate a ranking \cite{ckt:recsys2010}.
\end{itemize}

The objective function may be tinkered to handle certain of these
aspects. However, we think that the one and only way to really handle
the problem of recommendation is to address it as a sequential
decision making problem, since the history should be taken into
ccount. Such a sequential decision making problem faces an
exploration vs.\@ exploitation dilemma as detailed in section \ref{sec:bandits} 
 the exploration being meant to acquire information in
order to exploit it and to perform better subsequently; information
gathering has a cost that can not be merely minimized to 0, or simply
left as an unimportant matter.  This means that the evaluation of the
recommendation algorithm dealing with the cold start problem has to be
done online.

Based on these ideas, our contribution in this paper is the following:
\begin{quote}
we propose an original way to tackle the cold start problem of recommendation systems: we cast this problem as a sequential decision making problem to be played online that selects items to recommend in order to optimize the exploration/exploitation balance; our solution is then to perform the rating matrix factorization driven by the policy of this sequential decision problem in order to focus on the most useful terms of the factorization.

The reader familiar with the bandit framework can think of this work as a contextual bandit building its own context from the observed reward using the hypothesis of the existence of a latent space of dimension $k$.

We also introduce a methodology to use a classical partially filled rating matrix to assess the online performance of a bandit-based recommendation algorithm.
\end{quote}

After introducing our notation in the next section, Sec.\@
\ref{sec:a} presents the matrix factorization approach. Sec.\@
\ref{sec:bandits} introduces the necessary background in bandit
theory.
In Sec.\@ \ref{sec:cs-user} and Sec.\@ \ref{sec:cs-item}, we solve the
cold start setting in the case of new users and in the case of new
items. Sec.\@ \ref{sec:expe} provides an experimental study on
artificial data, and on real data. Finally, we conclude and draw some
future lines of work in Sec.\@ \ref{sec:future}.

\section{Notations and Vocabulary}
\label{sec:notations}

Uppercase, bold-face letters denote matrices, such as: $\mat{A}$.
$\mat{A}^T$ is the transpose matrix of $\mat{A}$, and $\mat{A}_i$
denotes its $i^{\mbox{\scriptsize{}th}}$ row $i$. Lowercase, bold-face
letters denote vectors, such as $\vect u$. $\# \vect u$ is the number
of components (dimension) of $\vect u$. Normal letters denote scalar
value.
Except for $\zeta$, greek letters are used to denote the parameters of
the algorithms.  We use calligraphic letters to denote sets, such as
${\cal S}$.  $\# {\cal S}$ is the number of elements of the set ${\cal
  S}$.  For a vector $\vect{u}$ and a set of integers ${\cal S}$
(s.t.\@ $\forall s \in {\cal S}, s\leq \# \vect{u}$), $\vect{u}_S$ is
the sub-vector of $\vect u$ composed of the elements of $\vect{u}$
which indices are contained in ${\cal S}$. Accordingly, $\mat U$ being
a matrix, ${\cal S}$ a set of integers smaller or equal to the number
of lines of $\mat U$, $\mat U_{\cal S}$ is the sub-matrix made of the
rows of $\mat U$ which indices form ${\cal S}$ (the ordering of the
elements in ${\cal S}$ does not matter, but one can assume that the
elements of the set are sorted). Now, we introduce a set of notations
dedicated to the \rs{} problem. We consider:

\begin{itemize}
  \item as we consider a time-evolving number of users and items, we will note $n$ the current number of users, and $m$ the current number of items. These should be indexed by a $t$ to denote time, though often in this paper, $t$ is dropped to simplify the notation. $i$ indices the users, whereas $j$
    indices the items. Without loss of generality, we assume $n < N$ and $m < M$, that is $N$ and $M$ are upper bounds of the number of ever seen users and items (those figures may as large as necessary).
  \item $\mat{R}^*$ represents the ground truth, that is the matrix of
    ratings. Obviously in a real application, this matrix is unknown.
    Each row is associated to one and only one user, whereas each
    column is associated to one and only one item. Hence, we will also
    use these row indices, and column indices to represent users, and
    items.

    $\mat{R}^*$ is of size $N\times{}M$. $r^*_{i,j}$ is the
    rating given by user $i$ to item $j$.

    We suppose that there exists an integer $k$ and two matrices
    $\mat{U}$ of size $N\times{}k$ and $\mat{V}$ of size $M\times{}k$
    such that $\mat{R}^* = \mat{U} \mat{V}^T$. This is a standard
    assumption \cite{Dror:2011fk}.

  \item Not all ratings have been observed. We denote $\knownRatings$
    the set of elements that have been observed (yet). Then we define
    $\mat{R}$:
    \[
      r_{i,j} = 
        \left\{ \begin{array}{l}
          r^*_{i,j} + \eta_{i,j} \mbox{ if } (i, j) \in \knownRatings\\
          \mbox{NA otherwise}
        \end{array} \right.
    \]
    where $\eta_{i,j}$ is a noise with zero mean, and finite
    variance. The $\eta_{i,j}$ are i.i.d.

    In practice, the vast majority of elements of $\mat{R}$ are unknown.

    In this paper, we assume that $\mat{R}^*$ is fixed during all the
    time; at a given moment, only a submatrix made of $n$ rows and $m$ columns is actually useful. This part of $\mat{R}^*$ that is observed is
    increasing along time. That is, the set $\knownRatings$ is
    growing along time.
  \item $\J(i)$ denotes the set of indices of the columns with
    available values in row number $i$ of $\mat R$ (\ie the set of
    items rated by user $i$). Likewise, ${\cal I}(j)$ denotes the sets
    of rows of $\mat{R}$ with available values for column $j$ (\ie the
    set of users who rated item $j$).
  \item Symbols $i$ and ${\cal I}$ are related to users, thus rows of
    the matrices containing ratings, while symbols $j$ and ${\cal J}$
    refer to items, thus columns of these matrices.
  \item $\hat{\mat{U}}$ and $\hat{\mat{V}}$ denote estimates (with the
    statistical meaning) of the matrices $\mat{U}$ and $\mat{V}$
    respectively. Their product $\hat{\mat{U}}\hat{\mat{V}}^T$ is denoted
    $\hat{\mat{R}}$. The relevant part of matrices $\mat R^*$ and $\mat R$ has dimensions $n_t \times m_t$ at a given moment $t$, $\mat U \in \mathbb{R}^{n_t \times{} k}$ and $\mat V \in \mathbb{R}^{m_t \times{} k}$.
\end{itemize}

To clarify things, let us consider $n = 4$ users, $m = 8$ items, and
$\mat{R}^*$ as follows:

\[
  \mat{R}^* = \left( \begin{array}{cccccccc}
    3 & 2 & 3 & 4 & 1 & 2 & 5 & 4 \\
    1 & 1 & 4 & 3 & 3 & 5 & 2 & 1 \\
    5 & 1 & 4 & 3 & 2 & 1 & 1 & 2 \\
    2 & 4 & 5 & 5 & 4 & 3 & 1 & 1 \\
  \end{array} \right)
\]

Let us suppose that $\knownRatings = \{ (1, 3), (1, 6), (2,
1), (2, 4), (2, 6), (3, 2), (4, 4), (4, 6), (4, 7) \}$, then 
assuming no noise:

\[
  \mat{R} = \left( \begin{array}{cccccccc}
    \mbox{NA} & \mbox{NA} & 3 & \mbox{NA} & \mbox{NA} & 2 & \mbox{NA} & \mbox{NA} \\
    1 & \mbox{NA} & \mbox{NA} & 3 & \mbox{NA} & 5 & \mbox{NA} & \mbox{NA} \\
    \mbox{NA} & 1 & \mbox{NA} & \mbox{NA} & \mbox{NA} & \mbox{NA} & \mbox{NA} & \mbox{NA} \\
    \mbox{NA} & \mbox{NA} & \mbox{NA} & 5 & \mbox{NA} & 3 & 1 & \mbox{NA} \\
  \end{array} \right)
\]


We use the term ``observation'' to mean a triplet (user, item, rating
of this item by this user such as $(2, 1, 1)$). Each known value of
$\mat{R}$ is an observation. The \rs{} receives a stream of
observations.  We use the term ``rating'' to mean the value associated
by a user to an item. It can be a rating as in the Netflix challenge,
or a no-click/click, no-sale/sale, \ldots
%
%

For the sake of legibility, in the online setting we omit the $t$
subscript for time dependency. In particular, ${\cal S}$, $\hat{\mat
  U}$, $\hat{\mat V}$, $n$, $m$ should be subscripted with $t$.

\section{Matrix Factorization}
\label{sec:a}

Since the Netflix challenge \cite{Bennett07thenetflix}, many works
have been using matrix factorization: the matrix of observed ratings
is assumed to be the product of two matrices of low rank $k$. We refer
the interested reader to \cite{Koren2009} for a short survey.
As most of the values of the rating matrix are unknown,
the factorization can only be done using this set of observed
values. The classical approach is to solve the regularized
minimization problem $(\hat{\mat{U}}, \hat{\mat{V}}) \stackrel{def}{=} \argmin{}_{\mat{U}, \mat{V}} \zeta(\mat{U}, \mat{V})$ where:

$$\zeta (\mat {U}, \vect{V}) \stackrel{def}{=} \sum_{\forall{}
  (i,j) \in {\cal S}} \left(r_{i,j} - \mat{U}_i\cdot \mat{V}_j^T \right)^2 +
{ \lambda \cdot \Omega (\mat{U}, \mat{V})},$$

in which $\lambda \in \mathbb{R}^+$ and the usual regularization term is:
$$ \Omega (\mat {U}, \mat {V}) \stackrel{def}{=} ||\mat U||^2+
||\mat V||^2 = \sum_i ||\mat U_i||^2 + \sum_j ||\mat V_j||^2.
$$

$\zeta$ is not convex. The
minimization is usually performed either by stochastic gradient
descent (SGD), or by alternate least squares (ALS).
\icmlVersion{ALS consists in iteratively solving for $\mat U$ while fixing $\mat V$ and then solve for $\mat V$ with this $\mat U$.}
\keepForLongVersionIfAny{Alternate least
squares consists in Alg. \ref{algo:ALS}. Steps 3 and 4 consists in
solving a convex optimization problem by the method of least-squares. Minimizing this function on a complete matrix with $\lambda=0$ is the same than computing a Singular Value Decomposition (SVD).

\begin{algorithm}
  \caption{Alternate Least Squares}
  \label{algo:ALS}
  \begin{algorithmic}[1]
    \REQUIRE 
    a matrix $\mat R$
    \STATE Initialize $\mat V$ with small random values, except the first row which is filled which $i^{\mbox{\tiny{}th}}$ element is the mean of the $i^{\mbox{\tiny{}th}}$ column of $\mat R$
    \FOR {$t=1,~2,\dots$}
      \STATE Compute $\mat U$ that minimizes $\zeta$ for a fixed $\mat V$
      \STATE Compute $\mat V$ that minimizes $\zeta$ for a fixed $\mat U$
    \ENDFOR
  \end{algorithmic}
\end{algorithm}
}
ALS-WR \cite{Zhou:2008:LPC:1424237.1424269} weighs users and items according to their respective importance in the matrix of ratings.

$$
 \Omega (\mat{U}, \mat{V}) \stackrel{def}{=} \sum_i \#\J(i) ||\mat U_i||^2 + \sum_j \# {\cal I}(j) ||\mat V_j||^2.
$$

This regularization is known to have a good empirical behavior ---
that is limited overfitting, easy tuning of $\lambda$ and $k$, low
RMSE.

\section{Bandits}
\label{sec:bandits}

Let us consider a bandit machine with $m$ independent arms. When
pulling arm $j$, the player receives a reward drawn from $[0,1]$ which
follows a probability distribution $\nu_j$. Let $\mu_j$ denote the
mean of $\nu_j$, $j^*\stackrel{def}{=}\argmax_j \mu_j$ be the best
arm and $\mu^*\stackrel{def}{=}\max_j \mu_j = \mu_{j^*}$ be the best expected reward (we assume there is only one best arm). The parameters $\{\nu_j\}$, $\{\mu_j\}$, $j^*$ and $\mu^*$ are unknown.

A player aims at maximizing its cumulative reward after $T$
consecutive pulls. More specifically, by denoting $j_t$ the arm pulled
at time $t$ and $r_t$ the reward obtained at time $t$, the player
wants to maximize the quantity $\cumrew_T=\sum_{t=1}^Tr_t.$ As the
parameters are unknown, at each time-step (except the last one), the
player faces the dilemma:
\begin{itemize}
  \item either \emph{exploit} by pulling the arm which seems the best
    according to the estimated values of the parameters;
 \item or \emph{explore} to improve the estimation of the parameters
   of the probability distribution of an arm by pulling it;
\end{itemize}
A well-known approach to handle the exploration vs.\@ exploitation
trade-off is the Upper Confidence Bound strategy (UCB)
\cite{Auer02finite-timeanalysis} which consists in
playing the arm $j_t$:
\begin{equation}\label{eq:UCB}
 j_t=\argmax_{j} \hat\mu_j + \sqrt{\frac{2\ln t}{t_j}},
\end{equation}
where $\hat\mu_j$ denotes the empirical mean reward incured when on
pulls of arm $j$ up to time $t$ and $t_j$ corresponds to the number of
pulls of arm $j$ since $t=1$.
UCB is optimal up to a constant. This equation clearly expresses
the exploration-exploitation trade-off: while the first term of the sum
($\hat\mu_j$) tends to exploit the seemingly optimal arm, the second
term of the sum tends to explore less pulled arms.

Li \etal \cite{LinUCB} extend the bandit setting to contextual
arms. They assume that a vector of real features $\vv \in \mathbb{R}^k$ is associated to each arm and that the expectation of the reward associated to an arm is $\uu^*\cdot\vv$, where $\uu^*$ is an unknown vector.
The algorithm handling this setting is known as LinUCB. LinUCB
follows the same scheme as UCB in the sense that it consists in playing
the arm with the largest upper confidence bound on the expected
reward:

$$j_t = \argmax_j \hat\uu.\vv_j^T + \alpha\sqrt{\vv_j\A^{-1}\vv_j^T},$$

where $\hat\uu$ is an estimate of $\uu^*$, $\alpha$ is a parameter and
$\A = \sum_{t'=1}^{t-1}\vv_{j_{t'}}.\vv_{j_{t'}}^T+\Id$, where
$\Id$ is the identity matrix. Note that $\hat\uu.\vv_j^T$
corresponds to an estimate of the expected reward, while
$\sqrt{\vv_j\A^{-1}\vv_j^T}$ is an optimistic correction of that
estimate.

While the objective of UCB and LinUCB is to maximize the cumulative
reward, theoretical results \cite{LinUCB,NIPS2011_1243} are expressed
in term of \emph{cumulative regret} (or regret for short)
$$\cumreg_T\stackrel{def}{=}\sum_{t=1}^Tr^*_t-r_t,$$ where $r^*_t$
stands for the best expected reward at time $t$ (either $\mu^*$ in the
UCB setting or $\max_j \uu^*.\vv_{j_t}^T$ in the LinUCB
setting). Hence, the regret measures how much the player looses (in
expectation), in comparison to playing the optimal strategy. Standard
results proove regrets of order $\tilde O(\sqrt T)$ or $O(\ln T)$,
depending on the assumptions on the distributions and depending on the
precise analysis
\footnote{$\tilde O$ means $O$ up to a logarithmic
  term on $T$.}.

Of course LinUCB, and more generally contextual bandits require the context (values of features) to be provided. In real applications this is done using side information about the items and the users \cite{Shivaswamy/Joachims/11b} --\ie expert knowledge, categorization of items, Facebook profiles of users, implicit feedback \ldots 
The core idea of this paper is to use matrix factorization techniques to build a context online using the known ratings. To this end, one assumes that the items and the arms can be represented in the same space of dimension $k$ and assuming that the rating of user $u$ for item $v$ is the scalar product of $u$ and $v$.

We study the introduction of new items and/or new users into the RS. This is done without using any side information on users or items.  

\section{Cold Start for a New User}
\label{sec:cs-user}

Let us now consider a particular recommendation scenario. 
At each time-step $t$,

\begin{enumerate}
  \item a user $i_t$ requests a recommendation to the \rs{},
  \item the \rs{} selects an item $j_t$ among the set of items that have
    never been recommended to user $i_t$ beforehand,
  \item user $i_t$ returns a rating $r_t=r_{i_t,j_t}$ for item $j_t$.
\end{enumerate}

Obviously, the objective of the \rs{} is to maximize the cumulative
reward $\cumrew_T=\sum_{t=1}^Tr_{t}$.

In the context of such a scenario, the usual matrix factorization
approach of \rs{} recommends item $j_t$ which has the best predicted
rating for user $i_t$. This corresponds to a pure exploitation
(greedy) strategy of bandits setting, which is well-known to be
suboptimal to manage $\cumrew_T$: to be optimal, the \rs{} has to
balance the exploitation and exploration.

\keepForLongVersionIfAny{
In fact, while the exploit strategy would maximize the expected
immediate reward $r_t$, this strategy prevents itself from discovering
promising items, and thus it induces a suboptimal cumulative
reward.

As a corollary, neither $r_t$, nor $\mat{R} - \hat{\mat{R}}$
measures the efficiency of the \rs{}. As is well-known in reinforcement
learning, the right measure of interest is the cumulative reward. 
In
the particular context of recommendation, any strategy leads to the
same limit cumulative reward as soon as $T\ge{}nm$. As a
consequence, to compare two \rs{}s, we have to consider their behavior
along time. More specifically, the faster they catch that limit value,
the better. An alternative measure of efficiency is the regret used in
bandit theory. This measure does not suffer from the drawback of
cumulative reward, as it sums up the difference between the optimal
solution at time $t$ and the selected one:
$$
\cumreg_T  \stackrel{def}{=} \sum_{t = 1}^{T} \max_{j \in \J_t(i_t) } r^*_{i,j} - r^*_{i,j_t}.
$$
}

Let us now describe the recommendation algorithm we propose at
time-step $t$. We aim at recommending to user $i_t$ an item $j_t$
which leads to the best trade-off between exploration and
exploitation in order to maximize  $\cumrew_\infty$. We assume that the matrix $\mat R$ is factorized into
$\hat{\mat U}\hat{\mat V}^T$ by ALS-WR - discussed later - which terminated by optimizing $\hat{\mat U}$ holding
$\hat{\mat V}$ fixed. In such a context, the UCB approach is based on
a confidence interval on the estimated ratings $\hat{r}_{i_t,j} =
\hat{\mat U}_{i_t}\cdot\hat{\mat V}_j^T$ for any allowed item $j$.

We assume that we already observed a sufficient number of ratings for
each item, but only a few ratings (possibly none) from user $i_t$. As a
consequence the uncertainty on $\hat{\mat U}_{i_t}$ is much more
important than on any $\hat{{\mat V}}_j$. In other words, the
uncertainty on $\hat{r}_{i_t,j}$ mostly comes from the uncertainty on
$\hat{\mat U}_{i_t}$. In the following, we express this uncertainty.

Let $\vect u^*$ denote the (unknown) true value of ${\mat U}_{i_t}$ and let us introduce the $k \times k$ matrix:

\begin{align*}
\mat A &\stackrel{def}{=} (\hat{\mat V}_{\J(i_t)})^T \cdot \hat{\mat V}_{\J(i_t)} + \lambda \cdot \#\J(i_t) \cdot \mat{Id}\\
&= \sum_{j \in \J(i_t)} \hat{\mat V}_j^T \cdot \hat{\mat V}_j + \lambda \cdot \#\J(i_t) \cdot \mat{Id}.
\end{align*}

As shown by \cite{Zhou:2008:LPC:1424237.1424269}, as $\hat{\mat U}$
and $\hat{\mat V}$ comes from ALS-WR (which last iteration optimized
$\hat{\mat U}$ with $\hat{\mat V}$ fixed),

\begin{align*}
\hat{ \mat U}_{j_t}&= \mat A^{-1} \sum_{j \in \J(i_t)} \hat{\mat V}_j^T \cdot r_{i_t,j}\\
&= \mat A^{-1} \hat{\mat V}_{\J(i_t)}^T {\mat R}_{i_t,\J(i_t)}^T.
\end{align*}

Using Azuma's inequality over the weighted sum of random variables (as
introduced by \cite{DBLP:journals/corr/abs-1205-2606} for linear
systems), it follows that there exists a value $C\in \R $ such as,
with probability $1 - \delta$:

$$
(\hat{\mat U}_{i_t} - \vect u^*) \mat{A}^{-1} (\hat{\mat U}_{i_t} - \vect u^*)^T  \leq C \frac{\log(1/\delta)}{t} 
$$

\begin{figure}
    \centering
    \begin{tikzpicture}[scale=0.7]
      \draw (0,-0.5) -- coordinate (x axis mid) (0,6);
      \draw (-0.5,0) -- coordinate (y axis mid) (6,0);
      \node at (-0.3,-0.3) {O};
      \node at (6,6) {$\mathbb{R}^k$};

      \draw[rotate = 40](3,2) ellipse (4 and 1)[color=blue, fill=blue, opacity=0.1];
	    \draw[rotate = 40](3,2) ellipse (4 and 1)[color=blue];
      \node[rotate = 40] at (0.8,3.4) {$\hat {\mat U}_{i_t}$ confidence ellipsoid};
    
      \shade[ball color=red] (3,2) circle (.1);
      \shade[ball color=red] (2,2) circle (.1); 
      \shade[ball color=red] (1,2.9) circle (.1);
      \shade[ball color=red] (2.5,3.2) circle (.1);     
      \shade[ball color=red] (4,3) circle (.1); 
      \shade[ball color=red] (5.5,3.2) circle (.1); 
      \node[color=red] at (5.45,2.8) {$\hat{\mat V}_2$};
      \node[color=red] at (3.4,1.7) {$\hat{\mat V}_1$};
      
      \draw[dashed] (0,0) -- coordinate (y axis mid) (6,4);
      \draw[dashed] (4.05,6.05) -- coordinate (y axis mid) (6.05,3);
      \shade[ball color=blue] (4.05,6.05) circle (.1); 
      \shade[ball color=blue] (5.6,3.7) circle (.1);
      \node[color=blue] at (5,3.85) {$\tilde {\vect u}^{(1)}$};

    \end{tikzpicture}
    \caption{This figure illustrates the use of the upper confidence
      ellipsoid for item selection in the context of a new user.
      As explained
      in the paper, items and users are represented as vectors in
      $\mathbb{R}^k$. In the figure, the red dots correspond to the known
      items vectors. The blue area indicates the confidence
      ellipsoid on the unknown vector associated to the user.
      The optimistic rating of the user for item $1$ is the maximum
      scalar product between $\hat{\mat V}_1$ and any point in this ellipsoid.
      By a simple geometrical argument based on
      iso-contours of the scalar product, this maximum scalar product is
      equal to the scalar product between $\hat{\mat V}_1$ and $\tilde {\vect u}^{(1)}$.
      The optimistic recommendation system recommends the item
      maximizing the scalar product $\langle\tilde {\vect u}^{(j)},~\hat{\mat V}_j\rangle$.}
    \label{fig:intuitionGeometrique}
\label{elip.user}
\end{figure}
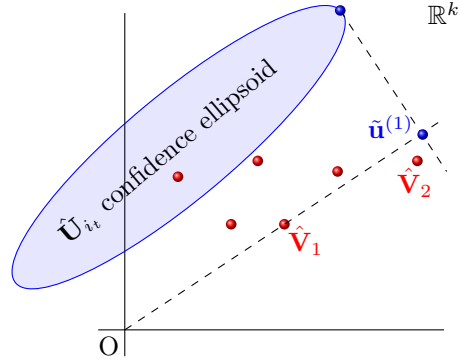

This inequality defines the confidence bound around the estimate $\hat{\mat U}_{i_t}$ of $\vect u^*$. Therefore, a UCB strategy selects item $j_t$:
$$
j_t \stackrel{def}{=} \argmax_{ 1\leq j \leq m, j \notin \J(i_t)} \;\; \max_{\vect u, s.t. ||\vect u - \hat{\mat U}_{i_t} ||^2_{A} < C \frac{\log(1/\delta)}{t}} \hat{\mat U}_{i_t} \cdot \hat{\mat V}_j 
$$

which amounts to:

$$
j_t = \argmax_{ 1\leq j \leq m, j \notin \J(i_t)} \hat{\mat U}_{i_t} \cdot \hat{\mat V}_j^T + \alpha \sqrt{\hat{\mat V}_j \mat A^{-1} \hat{\mat V}_j^T} 
$$

where $\alpha \in \R$ is an exploration parameter to be tuned.
Fig.\@ \ref{elip.user} illustrates the transition from the maximum on a
confidence ellipsoid to its closed-form $\hat{\mat U}_{i_t} \cdot \hat{\mat V}_j^T + \alpha \sqrt{\hat{\mat V}_j \mat A^{-1} \hat{\mat V}_j^T}$.

Our complete algorithm, named BeWARE.User (which stands for ``Bandit
WARms-up REcommenders'') is described in Alg.\@ \ref{BeWARE.User}. The
presentation is optimized for clarity rather than for computational
efficiency. Of course, if the exploration parameter $\alpha$ is set to
$0$ BeWARE.User chooses the same item as ALS-WR.
The estimate of the center of the ellipsoid and its size can be
influenced by the use of an other regularization term. BeWARE.User
uses a regularization based on ALS-WR. It is possible to replace all
$\#\J(.)$ by $1$. This amounts to the standard regularization: we call
this slightly different algorithm BeWARE.ALS.User. In fact one can use any regularization ensuring that $\hat {\mat U}_{i_t}$ is a linear combination of observed rewards. 
Please, note that BeWARE.ALS.User with $\lambda = 1$ is a LinUCB building its context using matrix decomposition - if the $\hat{\mat V}$ matrix does not changes after observation this is exactly a LinUCB.

\begin{algorithm}
  \caption{BeWARE.User: for a user $i_t$, selects and recommends an item to this user.
  }
  \label{BeWARE.User}
  \begin{algorithmic}[1]
    \REQUIRE 
      $i_t$,
      $\lambda$, 
      $\alpha$
    \ENSURE $\Rknown$, 
      ${\cal S}$
    \STATE ($\hat{\U}, \hat{\V}) \gets \MF(\Rknown)$
    \STATE $\A \gets (\hat{\mat V}_{\J(i_t)})^T \cdot \hat{\mat V}_{\J(i_t)} + \lambda \cdot \#\J(i_t) \cdot \mat{Id}$.
    \STATE\label{state:select.user}  $\displaystyle j_t \gets \argmax_{j \notin \J(i_t)} \hat{\mat U}_{i_t}\cdot\hat{{\mat V}}_{j}^T + \alpha\sqrt{\hat{{\mat V}}_j\A^{-1}\hat{{\mat V}}_j^T}$
    \STATE Recommend item $j_t$ and receive rating $r_t=r_{i_t,j_t}$
    \STATE Update $\Rknown$, ${\cal S}$
  \end{algorithmic}
\end{algorithm}


\subsection{Discussion on the Analysis of BeWARE.User}

The analysis of BeWARE.User is rather similar to the LinUCB proof
\cite{NIPS2011_1243} but it requires to take care of the vectors of
context which in our case are estimated through a matrix
decomposition. As matrix decomposition error bounds are classically
not distribution free \cite{Chatterjee:arXiv1212.1247} (they require
at least independancy between the observations), we cannot provide a
complete proof. However, we can have one for a modified algorithm
using the same LinUCB degradation as
\cite{journals/jmlr/ChuLRS11}. The trick is to inject some
independancy in the observed values in order to guarantee an unbiased
estimation of $V$.

\section{Cold Start for New Items}
\label{sec:cs-item}

When a new item is added, it is a larger source of uncertainty than
the descriptions of the users. To refect this fact, we compute a
confidence bound over the items instead of the users. As the second
step of the ALS is to fix $\hat{\mat U}$ and optimize $\hat{\mat V}$,
it is natural to adapt our algorithm to handle the uncertainty on
$\hat{\mat V}$ accordingly. This will take care of the exploration on
the occurrence of new items. With the same criterion and
regularization on $\hat{\mat V}$ as above, we obtain at timestep $t$:

$$
\mat B(j) \stackrel{def}{=} 
  (\hat{\mat U}_{{\cal I}(j)})^T \hat{\mat U}_{{\cal I}(j)} + 
    \lambda \cdot \# {\cal I}(j) \cdot \mat{Id} 
$$
and
$$
  \hat {\mat V}_j = 
    {\mat B(j)}^{-1}(\hat{\mat U}_{{\cal I}(j)})^T\mat R_{{\cal I}(j), j}.
$$

So considering the confidence ellipsoid on $\hat{\mat V}$, the upper
confidence bound of the rating for user $i$ on item $j$ is

$$
\hat{\mat U}_i \cdot \hat {\mat V}_j^T + \alpha \sqrt{\hat{\mat U}_j
\mat B(j)^{-1} \hat{\mat U}_j^T}.
$$

\begin{figure}
    \centering
    \begin{tikzpicture}[scale=0.7]
      \draw (0,-0.5) -- coordinate (x axis mid) (0,6);
      \draw (-0.5,0) -- coordinate (y axis mid) (6,0);
      \node at (-0.3,-0.3) {O};
      \node at (6,6) {$\mathbb{R}^k$};

      \shade[ball color=blue] (2.5,3) circle (.1);
	\node[color=blue] at (3,2.6) {$\hat{\mat U}_{i_t}$};
      
      \draw[rotate = 40](3,2) ellipse (1 and 0.2)[color=red,fill=red,opacity=0.1];
      \draw[rotate = 90](4,-5) ellipse (1 and 0.4)[color=red,fill=red,opacity=0.1];
      \draw[rotate = 10](1,2) ellipse (0.2 and 0.4)[color=red,fill=red,opacity=0.1];
      
      \draw[rotate = 0](1.5,3) ellipse (0.3 and 0.2)[color=red,fill=red,opacity=0.1];
      \draw[rotate = 30](0.5,2) ellipse (0.5 and 0.6)[color=red,fill=red,opacity=0.1];
      \draw[rotate = 0](3,0) ellipse (1.3 and 0.4)[color=red,fill=red,opacity=0.1];

      \draw[rotate = 40](3,2) ellipse (1 and 0.2)[color=red];
      \draw[rotate = 90](4,-5) ellipse (1 and 0.4)[color=red];
      \draw[rotate = 10](1,2) ellipse (0.2 and 0.4)[color=red];
      \draw[rotate = 0](1.5,3) ellipse (0.3 and 0.2)[color=red];
      \draw[rotate = 30](0.5,2) ellipse (0.5 and 0.6)[color=red];
      \draw[rotate = 0](3,0) ellipse (1.3 and 0.4)[color=red];

	\draw[rotate = 40](3,2) ellipse (1 and 0.2)[color=red];
	\draw[rotate = 90](4,-5) ellipse (1 and 0.4)[color=red];
	\draw[rotate = 10](1,2) ellipse (0.2 and 0.4)[color=red];
	\draw[rotate = 0](1.5,3) ellipse (0.3 and 0.2)[color=red];
	\draw[rotate = 30](0.5,2) ellipse (0.5 and 0.6)[color=red];
	\draw[rotate = 0](3,0) ellipse (1.3 and 0.4)[color=red];

	\node[color=red] at (5,4) {$\hat{\mat V}_j$};
        \draw[dashed] (0,0) -- coordinate (y axis mid) (5,6);
        \draw[dashed] (3.9,6.0) -- coordinate (y axis mid) (6,4.25);
	\shade[ball color=red] (5.15,4.95) circle (.1);

	\shade[ball color=blue] (4.55,5.45) circle (.1);
	\node[color=blue] at (3.9,5.4) {$\tilde \vv^{(j)}$};


    \end{tikzpicture}
    \caption{This figure illustrates the use of the upper confidence
      ellipsoid for item selection in the context of new items.  The
      setting is similar to Fig.\@ \ref{fig:intuitionGeometrique}
      except that the vector associated to the user is known (blue
      dot) while the items vectors live in confidence ellipsoids.  The
      optimistic recommendation system recommends the item maximizing
      the scalar product $\langle\hat{\mat U}_{i_t}, \tilde\vv^{(j)}\rangle$.}
\label{elip.item}
\end{figure}
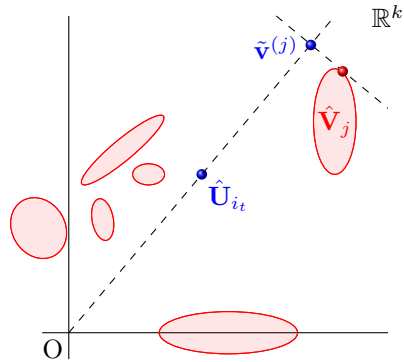

This leads to the algorithm BeWARE.Item presented in Alg.\@
\ref{BeWARE.Item}. Again, the presentation is optimized for clarity
rather than for computational efficiency. BeWARE.Item can be
parallelized and has the complexity of one step of ALS. Fig.\@
\ref{elip.item} gives the geometrical intuition leading to
BeWARE.Item. Again, setting $\alpha=0$ leads to the same selection as
ALS-WR. The regularization (on line 4) can be modified.
This algorithm has no straightforward interpretation in terms of
LinUCB.

\begin{algorithm}
  \caption{BeWARE.Item: for a user $i_t$, selects and recommends one of the new items to this user.
  }
  \label{BeWARE.Item}
  \begin{algorithmic}[1]
    \REQUIRE $i_t$,
      $\lambda$, 
      $\alpha$ 
    \ENSURE $\Rknown$, 
      ${\cal S}$ 
      \STATE ($\hat{\U}, \hat{\V}) \gets \MF(\Rknown)$
      \STATE $\forall j \notin \J(i_t),~ \mat B(j) \gets (\hat{\mat U}_{{\cal I}(j)})^T \hat{\mat U}_{{\cal I}(j)} 
       + \lambda \cdot \# {\cal I}(j) \cdot \mat{Id}$
      \STATE $\displaystyle j_t \gets 
        \argmax_{j \notin \J(i_t)}\hat{{\mat U}}_{i_t}.\hat{\mat V}_{j}^T + 
          \alpha\sqrt{\hat{\mat U}_{i_t}\B(j)^{-1}\hat{\mat U}_{i_t}^T}$
      \STATE Recommend item $j_t$ and receive rating $r_t=r_{i_t,j_t}$
      \STATE Update $\Rknown$, and ${\cal S}$
  \end{algorithmic}
  \label{step:updateB}
\end{algorithm}


\section{Experimental Investigation}
\label{sec:expe}


In this section we evaluate empirically our family of algorithms
on artificial data, and on real datasets. The BeWARE algorithms are
compared to:
\begin{itemize}
  \item greedy approaches (denoted \GreedyALS{} and
    \GreedyALSWR) that always choose the item with the
    largest current estimated value (respectively given a
    decomposition obtained by ALS, or by ALS-WR),
  \item the UCB1 approach \cite{Auer02finite-timeanalysis} (denoted
    \UCBonAllUsers{}) that consider each reward $r_{i_t,j_t}$
    as an independent realization of a distribution $\nu_{j_t}$. In
    other words, \UCBonAllUsers{} recommends an item without
    taking into account the information on the user requesting the
    recommendation.
\end{itemize}

On the one hand, the comparison to greedy approaches highlights the
needs of exploration to have an optimal algorithm in the online
context. On the other hand, the comparison to \UCBonAllUsers{} is there to assess the benefit of personalizing recommendations.

\begin{figure*}
  \begin{center}
    \subfigure[Artificial dataset.]
      { \includegraphics[width=0.3\linewidth]{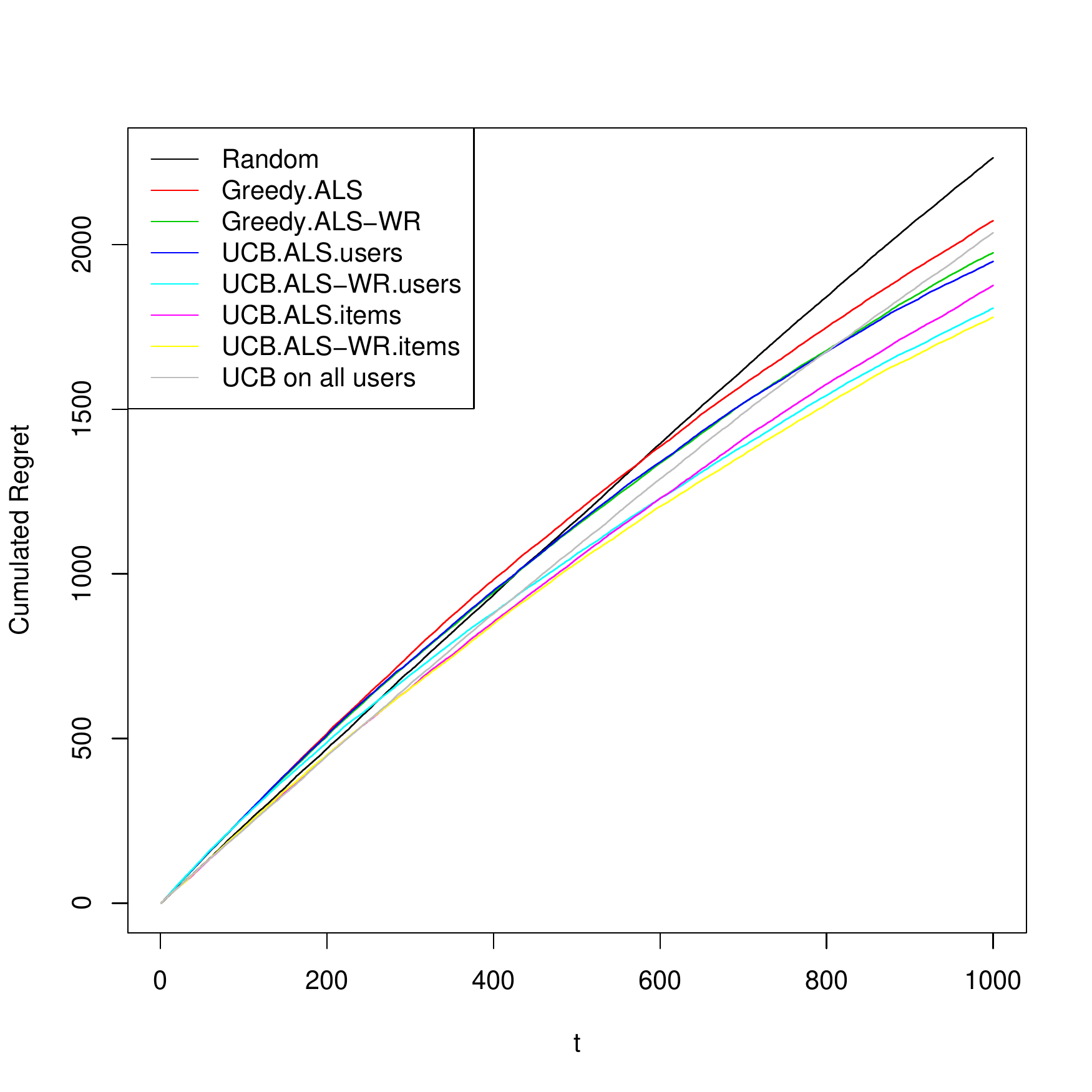}
        \label{RA}}
    \subfigure[Netflix dataset.]
      {\includegraphics[width=0.3\linewidth]{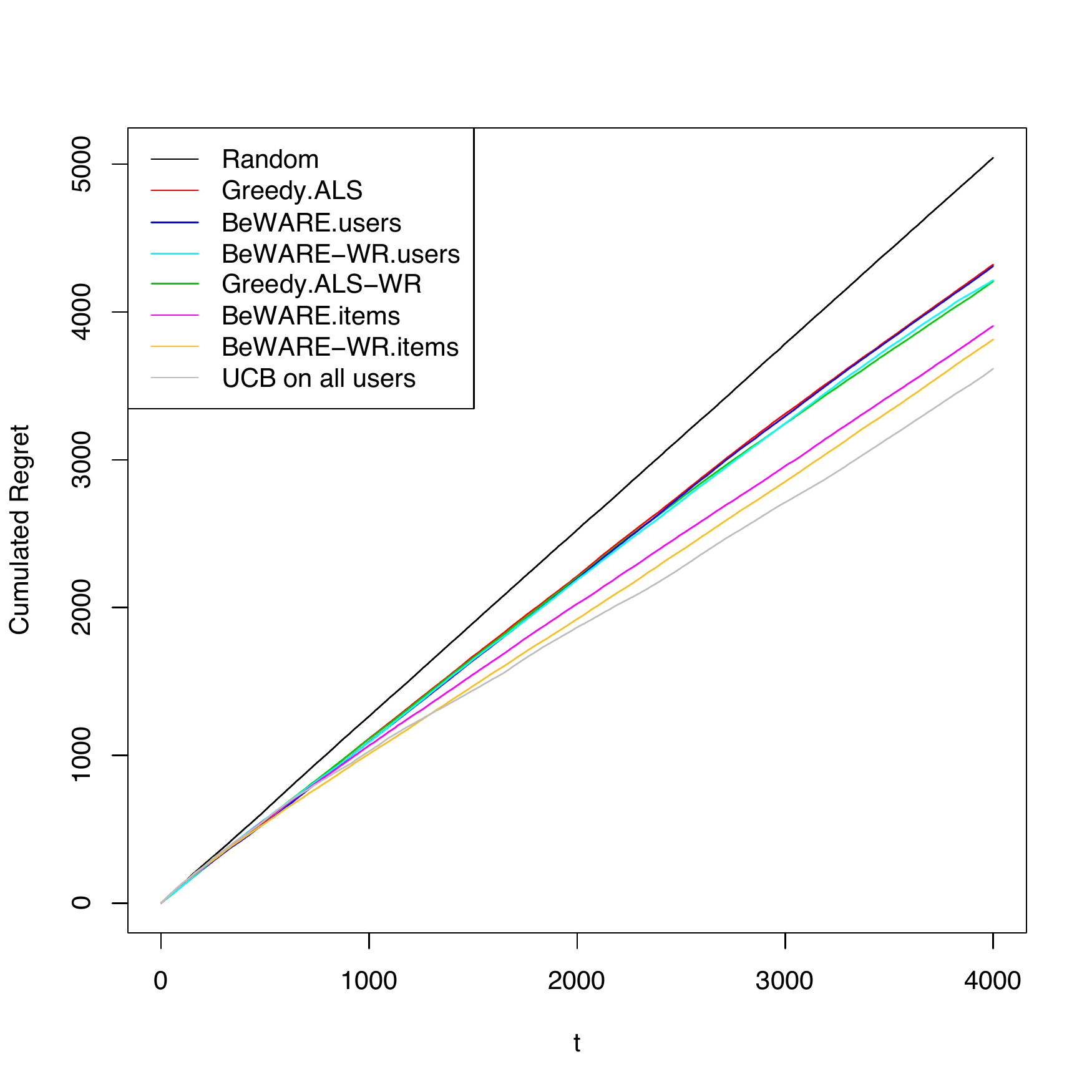}
        \label{RN}}
    \subfigure[Yahoo!Music dataset.]
      {\includegraphics[width=0.3\linewidth]{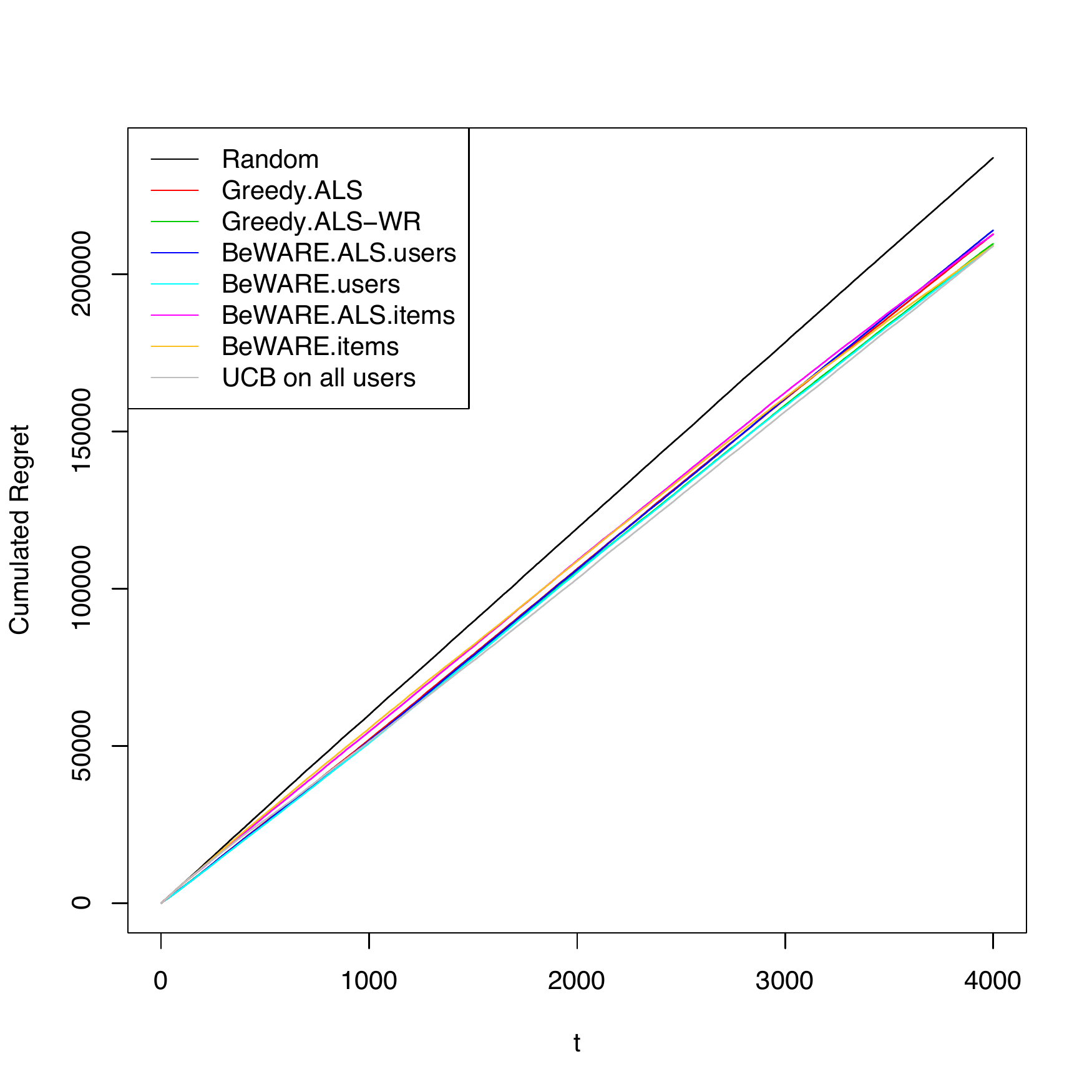}
        \label{RY}}
    \caption{Cumulated regret (the lower, the better) for a set of 100
      new items and 200 users with no prior information. Figures are
      averaged over $20$ runs (for Netflix and artificial data,
      $k=5,~\lambda=0.05,~ \alpha=0.12$ whereas for Yahoo!Music,
      $k=8,~\lambda=0.2,~ \alpha=0.05$). On the artificial dataset
      \subref{RA}, BeWARE.items is better than the other strategies in
      terms of regret. On the Netflix dataset \subref{RN}, \texttt{UCB
        on all users} is the best approach and BeWARE.items is the
      second best. On the Yahoo!Music dataset \subref{RY},
      BeWARE.items, \GreedyALSWR{} and UCB all 3 lead to similar
      performances.}
      \label{R}
  \end{center}
\end{figure*}

\subsection{Experimental Setting}

For each dataset, algorithms start with an empty $\Rknown$ matrix of
100 items and 200 users. Then, the evaluation goes like this:

\begin{enumerate}
  \item select a user uniformly at random among those who have not yet rated all the items,
  \item request his favorite item among those he has not yet rated,
  \item compute the immediate regret (the difference of rating between the best not yet selected item and the one selected according to $\widetilde\Rknown^*$ for this user),
  \item iterate until all users have rated all items.
\end{enumerate}

The difficulty with real datasets is that the ground truth is unknown, and actually, only a very small fraction of ratings is known. This makes the evaluation of algorithms uneasy. To overcome these difficulties, we also provide a comparison of the algorithms considering an artificial problem based on a ground truth matrix $\mat{R}^*$ considering $m$ users and $n$ items. This matrix is generated as in \cite{Chatterjee:arXiv1212.1247}. 
Each item belongs to either one of $k$ genres, and each user belongs
to either one of $l$ types. For each item $j$ of genre $a$ and each
user $i$ of type $b$, $r^*_{i,j}=p_{a,b}$ is the ground truth rating of
item $j$ by user $i$, where $p_{a,b}$ is drawn uniformly at random in
the set $\{1,2,3,4,5\}$. The observed rating $r_{i,j}$ is a noisy value
of $r^*_{i,j}$: $r_{i,j} = r^*_{i,j} + {\cal N}(0, 0.5)$.

We also consider real datasets, the NetFlix dataset
\cite{Bennett07thenetflix} and the Yahoo!Music dataset
\cite{Dror:2011fk}. Of course, the major issue with real data is that
there is no dataset with a complete matrix, which means we do no
longer have access to the ground truth $\mat {R^*}$, which makes the
evaluation of algorithms more complex. This issue is usually solved
in the bandit literature by using a method based on reject sampling
\cite{LiCLW11}. For a well constructed dataset, this kind of
estimators has no bias and a known bound on the decrease of the error
rate \cite{Langford_ExploScav_08}.

For all the algorithms, we restrict the possible choices for a user at
time-step $t$ to the items with a known rating in the dataset.
However, a minimum amount of ratings per user is needed to be able to
have a meaningful comparison of the algorithms (otherwise, a random
strategy is the only reasonable one). As a consequence, with both
datasets, we focus on the $5000$ heaviest users for the top ${\sim}
250$ movies/songs. This leads to a matrix $\widetilde\Rknown^*$ with
only $10\%$ to $20\%$ of missing ratings. We insist on the fact that
this is necessary for performance evaluation of the algorithms;
obviously, this is not required to use the algorithms on a live RS. 

For people used to work on full recommendation dataset the experiment can seem small. But one has to keep in mind several points:
\begin{itemize}
  \item Each value in the matrix corresponds to one possible observation. After each observation we are allowed to update our recommender policy. This means that for 4000 observations we need to perform 4000 matrix decompositions. 
  \item To evaluate precisely Beware, we would need the rating of any user on any item (because Beware may choose any of the items for the current user). In the dataset many of the ratings are unknown so using part of the matrix with many unknown ratings would introduce a bias in the evaluation. 
\end{itemize}  

We would like to advertize this experimental methodology which has a unique feature: indeed, this methodology allows us to turn any matrix --or tensor-- of ratings into an online problem which can be used to test bandit recommendation algorithms. 
This is of interest because there is currently no standard dataset to evaluate bandits algorithms. To be able to evaluate offline any bandit algorithm on real data, one has to collect data using a random uniform strategy and use a replay like methodology \cite{Langford_ExploScav_08}. To the best of our knowledge, the very few datasets with desired properties are provided by Yahoo Webscope program (R6 dataset) as used in the challenge \cite{ic12}.
These datasets are only available to academics which restrain their use. So, it is very interesting to be able to use a more generally available rating matrix (such as the Netflix dataset) to evaluate an online policy. We think that this methodology is an other contribution of this paper. A similar trick has already been used in reinforcement learning to turn a turn a reinforcement learning into a supervised classification task \cite{Lagoudakis03reinforcementlearning}. 

\subsection{Experimental Results}

Figures \ref{RA} and \ref{RN} show that given a fixed factorization
method, BeWARE strategies improve the results on the
Greedy-one. Looking more closely at the results, BeWARE based on items
uncertainty performs better than BeWARE based on users uncertainty,
and BeWARE.users is the only BeWARE strategy beaten by its greedy
counterpart (\GreedyALSWR) on the Netflix dataset. These
results demonstrate that an online strategy has to care about
exploration to tend towards optimality.

While \UCBonAllUsers{} is almost the worst approach over
Artificial data (Fig.\@ \ref{RA}), it surprisingly performs better
than all other approaches over Netflix dataset. We feel that this
difference is strongly related to the preprocessing of the Netflix
dataset we have done to be able to follow the experimental protocol
(and have an evaluation at all). By focusing on the top ${\sim} 250$
movies, we keep blockbusters that are appreciated by everyone. With
that particular subset of movies, there is no need to adapt the
recommendation user per user. As a consequence, \UCBonAllUsers{} suffers a smaller regret than other strategies, as it
considers users as $n$ independent realizations of the same
distribution. It is worth noting that \UCBonAllUsers{} regret
would increase with the number of items while the regret of BeWARE
scales with the dimensionality of the factorization, which makes
BeWARE a better candidates for real applications with much more items
to deal with.

Last, on Fig.\@ \ref{RY} all approaches suffer the same regret.

\subsection{Discussion}

In a real setting, BeWARE.Item has a desirable property: it
tends to favor new items with regards to older ones because they have less feedback
than the others, hence larger confidence bound. So the algorithm gives them a boost which is
exactly what a webstore is willing --- if a webstore accepts new
products this is because he feels the new one are potentially better
than the old ones. Moreover it will allow the recommender policy to
use at its best the novelty effect for the new items. This natural
attraction of users with regards to new items can be very strong as it
has been shown by the Exploration \& Exploitation challenge at
ICML'2012 which was won by a context free algorithm \cite{ic12}.

The computational cost of the BeWARE methods is the same as
doing an additional step of alternate least squares; moreover some
intermediate calculations of the QR factorization can be re-used to
speed up the computation. So the total cost of BeWARE.Item is
almost the same as ALS-WR.  Even better, while the online setting
requires to recompute the factorization at each time-step, this
factorization sightly changes from one iteration to the next
one. As a consequence, only a few ALS-WR iterations are needed to
update the factorization. Overall the computational cost stays
reasonable even in a real application.

\section{Conclusion and Future Work}
\label{sec:future}

In this paper, we introduced the idea of using bandit algorithm as a
principled, and effective way to solve the cold start problem in
recommendation systems. We think this contribution is conceptually
rich, and opens ways to many different studies. We showed on large,
publicly available datasets that this approach is also effective,
leading to efficient algorithms able to work online, under the
expected computational constraints of such systems. Furthermore, the
algorithms are quite easy to implement.

Many extensions are currently under study. First, we work on extending
these algorithms to use contextual information about users, and
items. This will require combining the similarity measure with
confidence bounds; this might be translated into a Bayesian prior. We
also want to analyze regret bound for large enough number of items and
users. This part can be tricky as LinUCB still does not have a
full formal analysis, though some insights are available in
\cite{NIPS2011_1243}.
 
An other important point is to work on the recommendation of several
items at once and get feedback only for the best one. There
is some work in the non contextual bandits on this point, one could
try to translate it in our framework
\cite{DBLP:journals/jcss/Cesa-BianchiL12}.

Finally, we plan to combine confidence ellipsoid about both users and
items --- this is not a straightforward sum of the bounds. However, we
feel that such a combination has low odds to provide better
results for real application, but it is interesting from a theoretical
perspective, and should lead to even better results on artificial
problems.


\end{document}